\documentclass{article} 

\usepackage{wrapfig}
\usepackage{microtype}
\usepackage{graphicx}
\usepackage{subcaption}
\usepackage{booktabs} 
\usepackage{multirow} 
\usepackage{hyperref}

\usepackage{iclr2027_conference,times}

\usepackage{amsmath}
\usepackage{amssymb}
\usepackage{mathtools}
\usepackage{amsthm}
\usepackage{enumitem}
\usepackage{mathtools}

\usepackage[capitalize,noabbrev]{cleveref}

\theoremstyle{plain}

\theoremstyle{definition}

\theoremstyle{remark}

\usepackage[textsize=tiny]{todonotes}

\usepackage{colortbl} 
\definecolor{mColor1}{rgb}{0.9,0.9,0.9}
\definecolor{mColor2}{rgb}{0.95,0.95,0.95}
\definecolor{non-photoblue}{rgb}{0.64, 0.87, 0.93}
\definecolor{lightblue}{rgb}{0.81, 0.94, 1.0}

\usepackage{amsmath,amsfonts,bm}

\def\eqref#1{equation~\ref{#1}}

\def\1{\bm{1}}

\DeclareMathAlphabet{\mathsfit}{\encodingdefault}{\sfdefault}{m}{sl}
\SetMathAlphabet{\mathsfit}{bold}{\encodingdefault}{\sfdefault}{bx}{n}

\usepackage{hyperref}
\usepackage{url}

\title{Instance-Adaptive Prompts as Context for Time-Series Foundation Models}

\author{%
\parbox[t]{\dimexpr\textwidth-2\tabcolsep\relax}{%
\centering
\normalfont
\textbf{%
Zehao Xiao\textsuperscript{1} \quad
Shifeng Xie\textsuperscript{1,2} \quad
Lei Zan\textsuperscript{1} \quad
Jianfeng Zhang\textsuperscript{3} \quad
Lujia Pan\textsuperscript{3}} \\[0.3em]
\textbf{%
Ievgen Redko\textsuperscript{1} \quad
Malik Tiomoko\textsuperscript{1} \quad
Keli Zhang\textsuperscript{1}}\\[0.8em]
\textsuperscript{1}Huawei Noah's Ark Lab, Paris, France\\
\textsuperscript{2}LIPADE, Universit\'e Paris Cit\'e, Paris, France\\
\textsuperscript{3}Huawei Noah's Ark Lab, Shenzhen, China
}%
}

\newcommand{\method}{{PaCTS}}

\iclrfinalcopy 
\begin{document}

\maketitle

\begin{abstract}
Longer histories can improve time-series foundation models (TSFMs), but require substantially higher inference cost. We therefore ask whether contextual information can be provided more efficiently through a compact set of learned token embeddings. We introduce PaCTS, which generates a small set of instance-adaptive latent prompts in the form of continuous embedding tokens conditioned on the visible context. These prompts serve as compact context surrogates for frozen TSFMs. PaCTS constructs them from instance-specific global statistics and further refines them with segment-level temporal information, capturing both global characteristics and local temporal variations. The prompt module is jointly trained and deployed across heterogeneous time series with the frozen backbone. Extensive experiments demonstrate the effectiveness of prompts as context, consistently improving forecasting across context lengths and model architectures. With a shorter input context, PaCTS can outperform the same frozen backbone using double context while requiring substantially less inference computation. Compared with weight-space adaptation methods, PaCTS achieves stronger improvements and better out-of-distribution generalization.
Code will be released at \url{https://github.com/zzzx1224/PaCTS}.
\end{abstract}

\section{Introduction}

\begin{wrapfigure}{r}{0.48\textwidth}
    \centering
    \vspace{-5mm}
    \includegraphics[width=\linewidth]{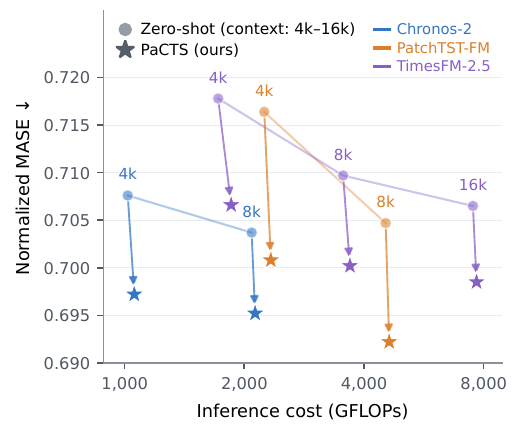}
    \vspace{-8mm}
    \caption{
        \textbf{\method~improves forecasting at lower additional cost than extending context.}
    }
    \vspace{-7mm}
    \label{fig:fig1}
\end{wrapfigure}
Foundation models (FMs) have achieved remarkable success in language \citep{achiam2023gpt,team2023gemini,liu2024deepseek} and vision tasks \citep{dosovitskiy2020image,radford2021learning}, demonstrating the ability of large-scale pretraining to deliver strong performance and generalization across tasks \citep{bommasani2021opportunities}.
Recently, time-series foundation models (TSFMs) have emerged as a general approach to address unique challenges of large-scale temporal data
\citep{woo2024moirai,auer2025tirex,podest2026tirex2,khwaja2026toto,xie2026tabby}.
Pretrained on diverse temporal data, these models achieve strong forecasting performance across datasets and domains without task-specific training \citep{ansari2025chronos2,das2024timesfm,meyer2026t0}.

TSFMs identify temporal patterns and forecast future values based on historical context.
Providing more history is therefore a natural way to supply additional context to the model.
However, processing longer histories increases inference cost,
even with patch-based tokenization
\citep{das2024timesfm,ansari2025chronos2,woo2024moirai}.
As Figure~\ref{fig:fig1} shows, increasing context length substantially
raises inference FLOPs across three TSFMs on GIFT-Eval, while yielding
only modest reductions in MASE.
Moreover, extending the context is not always an option. Many time series are too short to provide additional history
\citep{makridakis2018m4,aksu2024gift},
and sometimes longer inputs may exceed the context window supported by the pretrained backbone \citep{ansari2025chronos2,wen2026revisiting}.
We therefore ask:
\emph{can we provide a frozen TSFM with compact, informative context derived from the available history to improve forecasting?}

In this paper, we propose Prompts as Context for Time Series (\method), which provides compact, informative context to frozen TSFMs through instance-adaptive latent prompts.
These prompts are continuous embedding tokens adapted to each time series and prepended to the time-series patch tokens.
Our key perspective is to treat prompts as compact context, delivering forecasting benefits comparable to those of much longer histories through a learned summary of the input history, while keeping pretrained weights frozen.
\method~combines a shared, learned static prompt with an adaptive component generated from global statistics of the input.
The static component captures information shared across training series, while the adaptive component conditions the prompt on the current series.
Segment-level features further incorporate local temporal information into the prompts.
The entire prompt module is jointly trained on heterogeneous time series.
Once trained, it generates instance-adaptive prompts for new series without further parameter updates.
Although parameter-efficient, \method~is trained for shared use across datasets rather than separate adaptation to each target dataset.

We train and evaluate \method~on GIFT-Eval across multiple pretrained TSFM architectures and context lengths.
With trainable prompt parameters amounting to only less than 0.15\% of the corresponding pretrained model parameters and little additional inference computation, \method\ consistently reduces forecasting error, as shown in Figure~\ref{fig:fig1}.
Moreover, it achieves competitive and even better forecasting than doubling the historical context, with substantially lower inference cost.
By introducing adaptive prompts in the input space, \method~achieves better forecasting performance than weight-space adaptation alternatives and generalizes well to unseen distributions without further tuning.
These results demonstrate the effectiveness of learned prompts as context surrogates for time-series forecasting.

\section{Related Work}
\label{sec:related}

\textbf{Transformer-based time-series foundation models.}
Large-scale pretraining has enabled TSFMs to forecast across datasets
and domains without task-specific training
\citep{liang2024foundation}.
These models employ different representations of historical observations.
Chronos~\citep{ansari2024chronos} discretizes individual values into tokens,
whereas patch-based approaches group consecutive observations to reduce
sequence length.
Following the use of patching in PatchTST~\citep{nie2023time},
patch-based representations have been widely adopted by TSFMs,
including MOMENT~\citep{goswami2024moment},
Moirai~\citep{woo2024moirai},
TimesFM~\citep{das2024timesfm}, and Timer~\citep{liu2024timer}.
Recent models further explore mixture-of-experts architectures
\citep{liu2026timer} and hierarchical attention
\citep{sun2025xihe}.
Both decoder-only models, such as Moirai~2.0
\citep{liu2025moirai}, and encoder-only models, such as
Chronos-2~\citep{ansari2025chronos2} and
PatchTST-FM~\citep{wen2026revisiting}, achieve strong forecasting performance.
Our work builds on pretrained TSFMs and studies compact learned
context rather than changes to their backbone architectures.

\textbf{TSFM post-training.}
Pretrained forecasting models can be adapted through full fine-tuning or parameter-efficient updates \citep{xie2026post}.
LoRA~\citep{hu2022lora} learns low-rank weight updates, while selectively fine-tuning only updates subsets of pretrained parameters.
For TSFMs, Beyond LoRA~\citep{gupta2024beyond} studies BitFit, LayerNorm tuning, VeRA, and FourierFT.
Other work investigates multi-scale fine-tuning \citep{qiao2026multi} and compares full fine-tuning with LoRA across forecasting datasets~\citep{laglil2026foundation}.
Reinforcement-learning-based post-training has also been explored through feedback-driven policy optimization in TimeHF \citep{qi2025timehf} and forecasting-oriented reinforcement fine-tuning in TimeRFT~\citep{li2026timerft}.
GTN-R regularizes RL post-training by increasing predictive mass around training targets while encouraging diversity within their neighborhoods~\citep{zhang2026ground}.
Some other methods leave the backbone frozen.
TFMAdapter fits instance-level covariate-aware corrections \citep{dange2025tfmadapter}, while TS-Memory distills retrieval-based corrections into a parametric memory~\citep{lyu2026ts}.
Our approach keeps the pretrained backbone frozen and trains an efficient prompt module in input space.
Although parameter-efficient, it targets a shared prompt module trained across heterogeneous time series and reused without target-specific fine-tuning.

\textbf{Input-space conditioning.}
Prompt tuning~\citep{lester2021power} and prefix-tuning~\citep{li2021prefix} are first proposed for conditioning frozen language models through learned continuous prompts.
Visual prompt tuning~\citep{jia2022visual} extends prompts to vision Transformers. CoOp~\citep{zhou2022learning}, CoCoOp~\citep{zhou2022conditional} and the subsequent works \citep{khattak2023maple,xiao2024any} learn static and input-conditioned prompts for vision-language models.
Inspired by multimodal prompt tuning, UniCast~\citep{park2025unicast} incorporates visual and textual representations into time series prompts.  CoSPOT~\citep{choi2026compositional} combines spectral prompts with aligned time-series features in a frozen LLM for online forecasting.
Gen-P-Tuning~\citep{liu2025genpt} generates cross-channel prompts to adapt frozen univariate models to multivariate healthcare tasks.
In another direction, following the idea of in-context learning, TimesFM-ICF~\citep{faw2025icf} trains a forecaster to use related time-series examples supplied in context.
Motivated by context compression in LLM \citep{mu2023learning,chevalier2023adapting}, \method~directly learns latent prompts from the target time series as compact \emph{context surrogates} for frozen TSFMs.
The prompt module is trained across heterogeneous series and used without dataset-specific tuning, aiming to provide the forecasting benefit of longer histories at little additional inference cost.   

\section{Method}
\label{sec:method}

Let $f_\theta$ denote a pretrained TSFM with frozen parameters $\theta$. Given a univariate context $\mathbf{x}=(x_1,\dots,x_T) \in \mathbb{R}^{T}$ with $T$ time steps, the model first normalizes the input, partitions it into $L=\frac{T}{P}$ patches of size $P$, and maps them to token embeddings $\mathbf{E}= (\mathbf{e}_1,\dots,\mathbf{e}_L) \in \mathbb{R}^{L\times d}$, where $d$ is the model dimension. 
A Transformer backbone then produces a probabilistic forecast over the future $H$ time steps: 
\begin{equation}
\hat{\mathbf{y}}
=
f_\theta\big(\mathbf{E}(\mathbf{x})\big)
\in \mathbb{R}^{Q\times H},
\end{equation}
where $Q$ denotes the number of predicted quantiles.

For a frozen TSFM, extending the context $\mathbf{x}$ is the most direct way to expose the model to additional historical information. However, doing so lengthens the token sequence $\mathbf{E}$ and substantially increases inference cost, while the benefit of additional history is often limited. Moreover, longer contexts are not always available: many time series are inherently short \citep{makridakis2018m4,aksu2024gift}, and some backbones are trained with a finite context budget \citep{ansari2025chronos2,wen2026revisiting}, beyond which additional history may be discarded or even become harmful. We therefore ask whether forecasting-relevant information can instead be supplied through a second, more compact input channel.

\begin{figure}[t]
\centering
\includegraphics[width=\linewidth]{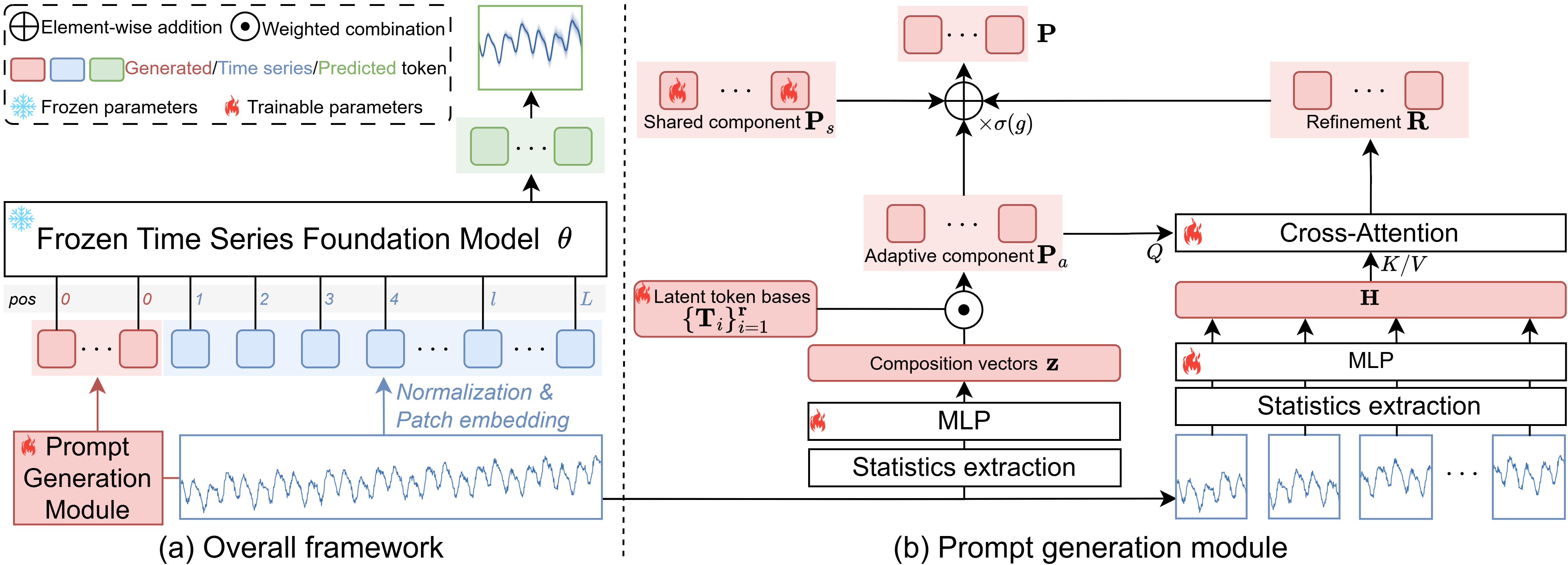}
\caption{\textbf{Overview of \method.}
(a) A lightweight prompt generation module maps the context to instance-adaptive latent prompts: continuous vectors prepended to the time-series embeddings.
(b) Global context statistics determine a weighted combination of learnable prompt basis patterns, yielding the adaptive component $\mathbf{P}_a(\mathbf{x})$.
Segment-level statistics are encoded with relative-position embeddings and retrieved through cross-attention to produce a refinement $\mathbf{R}(\mathbf{x})$.
The adaptive component and its refinement are combined, gated, and added to the shared component $\mathbf{P}_s$ to form the final prompts.
}
\label{fig:fig2}
\end{figure}

To this end, we propose \method, a lightweight post-training approach that supplies forecasting-relevant information through $M \ll L$ instance-adaptive latent prompt tokens, which are continuous embeddings generated from the visible context.
Prepended to the time-series embeddings, these prompts serve as compact context surrogates and improve forecasting without the cost of substantially extending the input context.
They are generated adaptively from the visible context and condition the frozen TSFM without modifying its pretrained parameters. Figure~\ref{fig:fig2} illustrates the overall framework.

\subsection{Latent Prompts as Context Surrogates}
\label{sec:method-prompt}

To construct compact context surrogates for the frozen TSFM, we introduce $M$ learnable embedding tokens $\mathbf{P}=(\mathbf{p}_1,\dots,\mathbf{p}_M)\in\mathbb{R}^{M\times d}$, which we refer to as \emph{latent prompts}. These prompts are continuous vectors in the time-series embedding space.
Following the prompt-tuning paradigm in language and vision models
\citep{lester2021power,zhou2022learning,zhou2022conditional},
we prepend them to the patch embeddings, yielding
\begin{equation}
\hat{\mathbf{y}} \;=\; f_\theta\big([\,\mathbf{P}\,;\,\mathbf{E}\,]\big),
\end{equation}
where $[\,\cdot\,;\,\cdot\,]$ denotes concatenation along the sequence dimension. 

During training, only the parameters associated with latent prompt generation are optimized, while all pretrained TSFM parameters $\theta$ remain frozen. 
The prompt tokens act as compact conditioning signals that supply forecasting-relevant information to the frozen model and thereby influence the predictions.
Since the prompts are prepended rather than substituted for existing tokens, the complete visible context is retained. 
With $M\!\ll\!L$, the latent prompts serve as efficient context surrogates with substantially lower computational overhead than extending the historical context.

\textbf{Position encoding.}
Transformer-based TSFMs use positional encodings to represent the temporal order of the input tokens. 
To preserve the original positional structure of the time-series tokens, we assign all prompt tokens with position index $0$, treating them as position-agnostic conditioning signals rather than additional temporal observations. The original time-series tokens retain their original positional indices.

\subsection{Instance-Adaptive Prompt Generation}
\label{sec:method-adaptive}

In its simplest form, $\mathbf{P}$ is a set of learnable latent prompt tokens shared across all inputs. Such a static prompt can capture regularities shared by the training corpus, but must accommodate heterogeneous time series with substantially different sampling frequencies, trends, seasonalities, and noise characteristics. A fixed prompt may therefore fail to capture instance-specific variations and overfit to training regularities, limiting its generalization to unseen series.
To address this limitation, we decompose the prompt into shared and input-adaptive components:
\begin{equation}
\mathbf{P}(\mathbf{x}) \;=\; \underbrace{\mathbf{P}_{s}}_{\text{shared}}
\;+\; \sigma(g) \cdot \underbrace{\mathbf{P}_{a}(\mathbf{x})}_{\mathclap{\text{sample-adaptive}}},
\end{equation}
where $\sigma(g)$ is a sigmoid gate with a trainable scalar $g$ controlling the magnitude of the adaptive component.
The shared component $\mathbf{P}_{s}\in\mathbb{R}^{M\times d}$ is a free parameter matrix learned across inputs and initialized using the mean patch embedding of the training data, placing the initial prompts near typical time-series token representations.
The adaptive component $\mathbf{P}_{a}(\mathbf{x})$ is input-dependent and tailors the prompts to the characteristics of each input series.

To generate $\mathbf{P}_{a}(\mathbf{x})$, we extract a compact feature vector $\mathbf{s}\in\mathbb{R}^{N_s}$ from the input series, comprising scale-invariant statistics that characterize its distributional and temporal structure, including trend, variability, autocorrelation, and noise level. 
Their definitions are provided in Appendix~\ref{sec:stats}.
All statistics are computed non-parametrically in normalized space, ensuring invariance to the location and scale of the input series.

Since $\mathbf{s}$ is low-dimensional, directly predicting all ${M\times d}$ entries of the adaptive prompt would require a large output layer. 
We therefore factorize the generation through a {latent prompt basis} consisting of $r$ learnable patterns $\{\mathbf{T}_1,\dots,\mathbf{T}_r\}\subset\mathbb{R}^{M\times d}$:
\begin{equation}
\mathbf{P}_{a}(\mathbf{x}) \;=\; \sum_{i=1}^{r} z_i\cdot\mathbf{T}_i,
\qquad \mathbf{z}=\mathrm{MLP} (\mathbf{s}) \in \mathbb{R}^{r},
\end{equation}
where a small MLP maps the statistics to a composition vector $\mathbf{z}$ that determines how the basis patterns are combined.
This parameterization keeps the generator compact and confines instance-dependent variation to a learned low-dimensional prompt subspace.

The compact statistics capture forecasting-relevant distributional and temporal characteristics of the input series with little computational overhead, as they are computed non-parametrically and remain low-dimensional. 
Integrating instance-specific statistics into prompt generation further enables the prompt to adapt to each series at inference time, promoting generalization across heterogeneous inputs and mitigating overfitting to training data.

\subsection{Segment-Level Prompt Refinement}
\label{sec:method-segment}

The global statistics used to generate $\mathbf{P}_{a}(\mathbf{x})$ summarize the entire context. 
However, they may overlook local changes that matter for forecasting, such as recent shifts in trend or variability, particularly for long contexts.
To capture such temporal variation while keeping the prompts compact, we refine the adaptive prompts using segment-level information.

To do so, we extract up to $S$ contiguous, non-overlapping segments from the visible context, working backward from the most recent observation. 
The segment length adapts to the available history, subject to a minimum length for reliable statistics.
Shorter histories therefore yield fewer segments. 
For each segment, we compute the same set of statistics used by the global prompt generator on the normalized series and encode them with a small MLP. 
We further add learned relative-position embeddings to distinguish earlier and more recent segments. 
This yields a $d_a$-dimensional descriptor for each segment.
Stacking these descriptors forms the sequence $\mathbf{H}(\mathbf{x})\in\mathbb{R}^{S\times d_a}$, with unused segment slots masked out.

The adaptive prompts then retrieve relevant local information through multi-head cross-attention:
\begin{equation}
\mathbf{R}(\mathbf{x})
=
\operatorname{CrossAttn}\big(
\mathbf{P}_{a}(\mathbf{x}),\,
\mathbf{H}(\mathbf{x}),\,
\mathbf{H}(\mathbf{x})
\big),
\label{eq:segment-refinement}
\end{equation}
where $\mathbf{P}_{a}(\mathbf{x})$ are projected into queries and $\mathbf{H}(\mathbf{x})$ are projected into keys and values.
The attention module projects its output back to the model dimension, yielding $\mathbf{R}(\mathbf{x})\in\mathbb{R}^{M\times d}$.
We add this refinement to the adaptive component to obtain the final prompts:
\begin{equation}
\mathbf{P}(\mathbf{x}) = \mathbf{P}_{s} + \sigma(g)\cdot \big[\mathbf{P}_{a}(\mathbf{x}) + \mathbf{R}(\mathbf{x}) \big].
\label{eq:refined-prompt}
\end{equation}
The output projection of the cross-attention module is initialized to zero, so the refinement initially leaves the global adaptive prompts unchanged.
For inputs with at most one valid segment, we set
$\mathbf{R}(\mathbf{x})=\mathbf{0}$, retaining the global adaptive prompts.

This refinement allows the prompts to combine instance-level characteristics with temporally localized information without increasing the number of prompt tokens. The same compact prompt configuration can therefore accommodate histories of different lengths, supporting joint training across heterogeneous series.

\subsection{Training}
\label{sec:method-training}

We train \method~jointly across heterogeneous time-series datasets while keeping the pretrained TSFM frozen.
For quantile forecasting, we optimize the pinball loss over each sample's valid forecast positions:
\begin{equation}
\mathcal{L}
=
\frac{1}{Q|\mathcal{H}|}
\sum_{q=1}^{Q}
\sum_{t\in\mathcal{H}}
\rho_{\tau_q}\big(
\tilde{y}_t-\hat{\tilde{y}}_t^{(q)}
\big),
\label{eq:training-loss}
\end{equation}
where $\tau_q$ denotes the $q$-th quantile level,
$\rho_{\tau}(u)=\max(\tau u,(\tau-1)u)$,
and $\mathcal{H}$ contains the valid forecast positions of the sample.
The targets $\tilde{y}_t$ and predictions
$\hat{\tilde{y}}_t^{(q)}$ are expressed in the backbone's normalized space.

Training samples may have different forecast horizons or missing target values.
We therefore mask invalid and padded positions and normalize each sample's loss by its own number of valid positions.
This prevents samples with shorter valid horizons from being down-weighted by padding.
The training objective is averaged over samples in each batch.

Only the parameters of the shared prompts, the instance-adaptive generator, and the segment-refinement module are optimized.
All pretrained backbone parameters $\theta$ remain frozen.
For each backbone, a single prompt module is trained across datasets and generates instance-adaptive latent prompts at inference time without further parameter updates.

\section{Experiments}
\label{sec:experiments}

\subsection{Experimental Setup}
\label{sec:exp-setup}

\textbf{Benchmarks and data splits.}
We evaluate \method~on two time series benchmarks, GIFT-Eval \citep{aksu2024gift} and TIME \citep{qiao2026s}.
GIFT-Eval serves as our primary benchmark, consisting of 35 datasets with 97 configurations defined by task and forecast horizon.
Training, validation, and test sets follow the standard time-based splitting of GIFT-Eval: the first 80\% of each series for training, the next 10\% for validation, and the final 10\% for testing, preserving temporal ordering and preventing data leakage.
We train a single prompt module jointly across all datasets and evaluate it across all test sets.
To assess generalization ability, we evaluate the GIFT-trained module directly on TIME, which contains 98 forecasting tasks from 50 datasets with no dataset overlap with GIFT-Eval, without further tuning.
Dataset configuration and split construction are provided in Appendix~\ref{app:data}.

\textbf{Backbones and baselines.}
We use Chronos-2 \citep{ansari2025chronos2} as the primary backbone and further evaluate \method~on PatchTST-FM \citep{wen2026revisiting} and TimesFM-2.5 \citep{das2024timesfm} to examine its applicability across encoder-only and decoder-only models.

To assess the effectiveness of \method~as a post-training approach, we compare it on Chronos-2 against the zero-shot backbone and five adaptation baselines: full fine-tuning, LoRA \citep{hu2022lora}, LayerNorm-only tuning \citep{zhao2024tuning}, BitFit \citep{zaken2022bitfit}, and linear probing \citep{kumar2022fine}.
Full fine-tuning and LoRA follow the official Chronos-2 training recipes. The remaining baselines use the same training pipeline while updating only their respective parameter subsets.
For other backbones, we compare \method~with the zero-shot model under the same evaluation protocol.

\textbf{Evaluation protocol and metrics.}
For evaluation, we measure point and probabilistic forecasting accuracy using mean absolute scaled error (MASE) and continuous ranked probability score (CRPS), respectively.
Following the benchmark protocol, we normalize scores by the corresponding seasonal-naive baseline and aggregate across forecasting configurations using the geometric mean.
We train the prompt module in a univariate setting and evaluate \method~and the baselines under the univariate protocol.
We additionally evaluate the same module under the multivariate protocol without retraining and report the results separately.
For inference efficiency, we measure FLOPs, GPU memory, and latency compared to the zero-shot baseline.

\textbf{Training and implementation.}
We keep the backbone frozen and jointly train a single prompt module across datasets for each backbone and context-length setting.
Our default configuration uses $M=10$ latent prompt tokens, a rank $r=4$ adaptive generator, and up to $S=16$ segments for prompt refinement.
We optimize the training parameters by AdamW with learning rate $0.001$ and select checkpoints using validation loss.
Each trained module is shared across evaluation tasks without task-specific adaptation.
More details are in Appendix~\ref{app:training}. All codes will be available.

\subsection{Results}
\label{sec:exp-results}

\begin{table*}[t]
    \centering
    \caption{
        Forecasting performance on GIFT-Eval across 97 configurations under the univariate protocol based on Chronos-2.
        Adaptation results are averaged over three seeds.
        \method~achieves best performance  across both context lengths and metrics, even lower MASE and CRPS than the zero-shot and five adaptation baselines with twice the context length (\method~4,096 vs. others 8,192).
    }
    \label{tab:main-results}
    \setlength{\tabcolsep}{6pt}
    \renewcommand{\arraystretch}{1.0}
    \resizebox{\columnwidth}{!}{%
    \begin{tabular}{lrrrrrrrr}
        \toprule
        & \multicolumn{4}{c}{Context length 4,096}
        & \multicolumn{4}{c}{Context length 8,192} \\
        \cmidrule(lr){2-5}
        \cmidrule(lr){6-9}
        Method
        & MASE $\downarrow$ & $\Delta$ (\%)
        & CRPS $\downarrow$ & $\Delta$ (\%)
        & MASE $\downarrow$ & $\Delta$ (\%)
        & CRPS $\downarrow$ & $\Delta$ (\%) \\
        \midrule
        Chronos-2 zero-shot
        & 0.708 & --- & 0.496 & ---
        & 0.704 & --- & 0.491 & --- \\
        \midrule
        Linear probing
        & 0.710 & +0.28 & 0.491 & -1.01
        & 0.705 & +0.14 & 0.485 & -1.22 \\
        BitFit
        & 0.708 & 0.00 & 0.489 & -1.41
        & 0.704 & 0.00 & 0.485 & -1.22 \\
        LayerNorm-only
        & 0.705 & -0.42 & 0.488 & -1.61
        & 0.701 & -0.43 & 0.483 & -1.63 \\
        Full fine-tuning
        & 0.703 & -0.71 & 0.487 & -1.81
        & 0.698 & -0.85 & 0.482 & -1.83 \\
        LoRA
        & 0.703 & -0.71 & 0.493 & -0.60
        & 0.699 & -0.71 & 0.488 & -0.61 \\
        \midrule
        \rowcolor{lightblue}
        \method
        & \textbf{0.697} & \textbf{-1.55} & \textbf{0.480} & \textbf{-3.23}
        & \textbf{0.695} & \textbf{-1.28} & \textbf{0.478} & \textbf{-2.65} \\
        \bottomrule
    \end{tabular}}
\end{table*}

\textbf{Main forecasting results.}
Table~\ref{tab:main-results} compares \method~with the zero-shot Chronos-2 backbone and five adaptation baselines on GIFT-Eval.
At both context lengths of 4,096 and 8,192, \method~improves over the zero-shot backbone on both MASE and CRPS and achieves the lowest average scores among the compared methods.
Full fine-tuning and LoRA improve both metrics at both context lengths, but their gains are smaller than those of \method.
The other methods primarily improve CRPS, with more limited or no gains in MASE.
These results demonstrate the effectiveness of \method~as a post-training approach, improving both point and probabilistic forecasting while keeping the backbone frozen.


\begin{table}[t]
    \centering
    \caption{
        \textbf{Forecasting performance and inference costs of Chronos-2 on GIFT-Eval.}    
        Compared with the pretrained model at doubled context length, our method achieves better forecasting performance at lower inference cost, with only 0.128\% additional parameters, demonstrating the effectiveness of latent prompts as context surrogates.
    }
    \label{tab:efficiency}
    \renewcommand{\arraystretch}{1.12}
    \resizebox{\columnwidth}{!}{%
        \begin{tabular}{lrrrrrr}
            \toprule
            Method (context)
            & MASE $\downarrow$
            & CRPS $\downarrow$
            & GFLOPs $\downarrow$
            & Memory (MiB) $\downarrow$
            & Latency (ms) $\downarrow$
            & Params (M) \\
            \midrule
            Zero-shot (4,096)
            & 0.708 & 0.496 & 1,018.4 & 651.3 & 40.4 & 119.478 \\
            Zero-shot (8,192)
            & 0.704 & 0.491 & 2,087.4 & 809.4 & 75.0 & 119.478 \\
            \rowcolor{lightblue}
            \method~(4,096)
            & \textbf{0.697} & \textbf{0.480}
            & 1,057.9 & 660.7 & 47.4 & 119.631 \\
            \midrule
            $\Delta$ vs.\ zero-shot (8,192)
            & $-0.99\%$ & $-2.24\%$
            & $-49.32\%$ & $-18.37\%$ & $-36.80\%$ & $+0.128\%$ \\
            \bottomrule
        \end{tabular}%
    }
\end{table}

\textbf{Prompts as context surrogates.}
We next examine whether \method~can match longer-context forecasting at a lower inference cost.
Table~\ref{tab:main-results} shows that \method~at 4,096 steps achieves lower mean MASE and CRPS than the zero-shot backbone and all five adaptation baselines using twice the context length, 8,192 steps.

Table~\ref{tab:efficiency} further reports inference costs for inputs that fill the context window.
Compared with zero-shot inference at 8,192 steps, \method~at 4,096 steps reduces FLOPs by 49.3\%, peak GPU memory by 18.4\%, and latency by 36.8\%.
At the same context length of 4,096, prompt generation and processing add only 3.9\% FLOPs and 1.4\% memory, with a latency overhead of 17.3\%.
The prompt module adds only 153,286 trainable parameters, increasing the total model size by 0.128\%.
These results support the effectiveness of latent prompts as context surrogates, achieving competitive and even better average forecasting accuracy with a shorter visible history and lower inference cost.

\begin{wraptable}{r}{0.48\textwidth}
    \centering
    \vspace{-4mm}
    \caption{
        \textbf{Ablations on prompt generation.}
        Prompt module is tuned based on Chronos-2 at 4,096 steps. Both adaptive generation and segment refinement improves forecasting.
    }
    \vspace{-2mm}
    \label{tab:ablation}
    \small
    \setlength{\tabcolsep}{3pt}
    \renewcommand{\arraystretch}{1.12}
    \begin{tabular*}{\linewidth}{@{\extracolsep{\fill}}lrr@{}}
        \toprule
        Variant & MASE $\downarrow$ & CRPS $\downarrow$ \\
        \midrule
        Zero-shot             & 0.708 & 0.496 \\
        \midrule
        Static prompts        & 0.707 & 0.490 \\
        + Adaptive generation & 0.704 & 0.486 \\
        + Segment refinement  & \textbf{0.697}
                              & \textbf{0.480} \\
        \bottomrule
    \end{tabular*}
    \vspace{-6mm}
\end{wraptable}
\textbf{Ablations on prompt generation.}
We also evaluate the contributions of instance-adaptive prompt generation and segment-level refinement using Chronos-2 at a context length of 4,096.
As shown in Table~\ref{tab:ablation}, static prompts improve CRPS but provide little improvement in MASE.
Conditioning the prompts on the input improves both metrics, and adding segment-level refinement yields further gains.
These results support the complementary roles of instance-level conditioning and local temporal information in constructing adaptive and effective context surrogates.


\textbf{Effects of history length and forecast horizon.}
We further examine how the gains of \method~vary across history lengths and forecast horizons on GIFT-Eval.
We evaluate \method, LoRA, and full fine-tuning with a context length of 4,096 steps on GIFT-Eval configurations grouped separately by history length and forecasting horizon, and compare them with pretrained Chronos-2 with the same and double context length.

\begin{figure}[t]
    \centering
    \includegraphics[width=\textwidth]{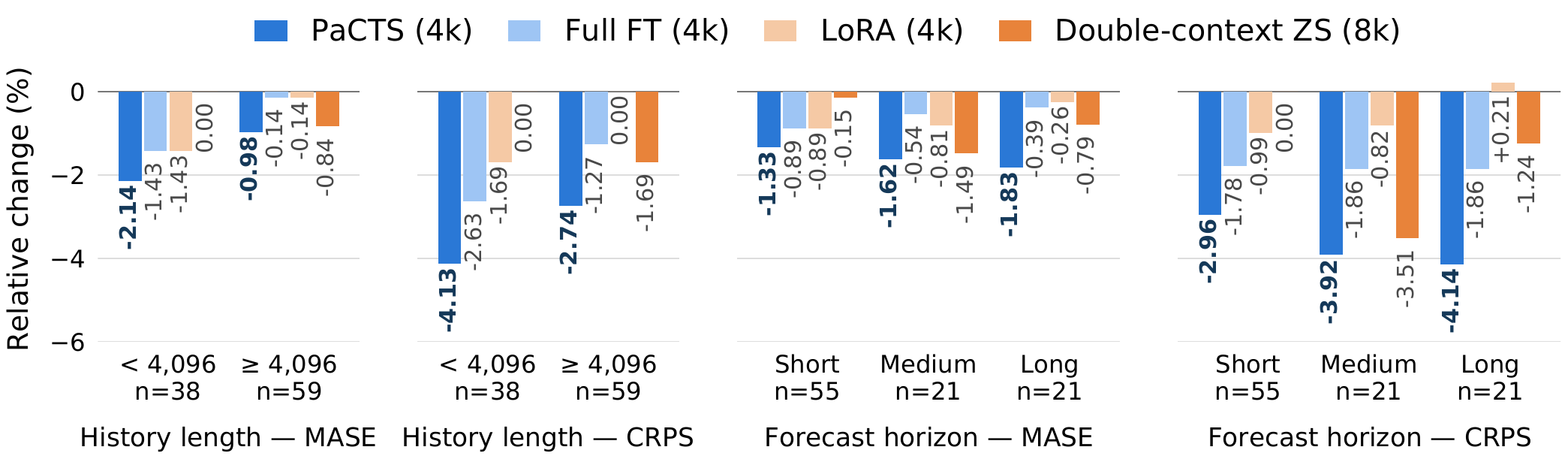}
    \caption{
        \textbf{Effects of history length and forecast horizon.} The bars show percentage changes in MASE and CRPS relative to zero-shot Chronos-2 at 4,096 steps. Negative values indicate improvements.
        The left two panels group configurations by history lengths, and the right two by forecasting horizons.
        \method~performs consistently better than other methods and double-context zero-shot forecasts. Detailed numbers are in Appendix~\ref{app:exp}.
    }
    \label{fig:history-horizon}
\end{figure}

As shown in the left two panels of Figure~\ref{fig:history-horizon}, doubling the context length improves performance primarily in the longer-history group ($\geq 4,096$ steps), while providing no improvement in the shorter-history group ($< 4,096$ steps) since the available history is already covered by the original context window.
LoRA and full fine-tuning also exhibit distinct behavior across these groups. Their gains are larger on shorter histories, with limited MASE improvements on longer histories.
In contrast, \method~improves both metrics and outperforms zero-shot inference with double the context length in both groups, supporting the effectiveness of latent prompts as context surrogates.

Across forecasting horizons, the MASE gains of LoRA and full fine-tuning diminish in the long-horizon group, and LoRA even yields worse CRPS than the pretrained baseline.
Doubling the context length provides good gains for medium and long horizons, but only marginal gains for short ones.
In contrast, \method~outperforms both adaptation baselines and double-context inference on both metrics across all three horizon groups, which demonstrates consistent benefits across different forecasting lengths.

\begin{figure}
    \centering
    \includegraphics[width=\linewidth]{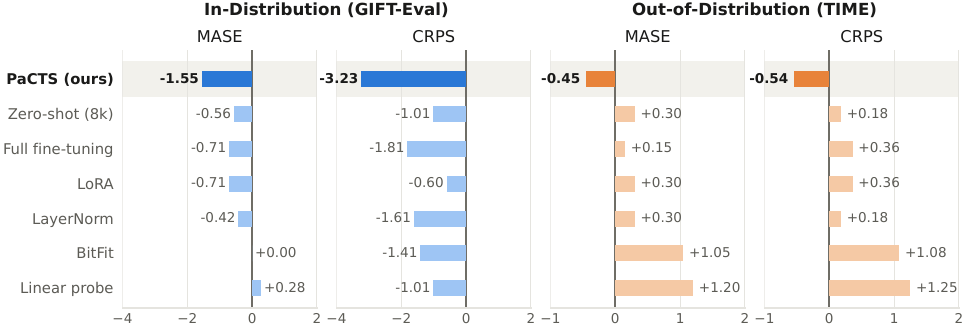}
    \caption{
        \textbf{In-distribution and out-of-distribution generalization performance comparisons.} 
        Relative changes in MASE and CRPS are measured against zero-shot Chronos-2 at 4,096 steps. Among the compared methods, \method~achieves the best in-distribution performance on GIFT-Eval and is the only method that improves both metrics over zero-shot inference on the TIME benchmark.
    }
    \label{fig:time-transfer}
\end{figure}
\textbf{Generalization to unseen data.}
We next evaluate the generalization ability of \method~to unseen data after adaptation.
We directly apply the prompt module trained on GIFT-Eval to TIME without further training or tuning, with 4,096 steps context length.
As shown in Figure~\ref{fig:time-transfer}, \method~is the only method that achieves lower MASE and CRPS on TIME than the original pretrained Chronos-2.
In contrast, all five weight-space adaptation baselines yield higher errors on both metrics, indicating that their improvements on GIFT-Eval do not carry over to TIME.
Notably, increasing context length of the original pretrained model to 8,192 also fails to improve either metric.
The results demonstrate that the proposed \textit{prompts as context} yields good generalization ability.

\begin{table}[t]
    \centering
    \caption{
        \textbf{Extension to multivariate forecasting with univariate-trained prompts.}
        \method~is trained only on the GIFT-Eval training split under the univariate setting and evaluated directly using the official multivariate inference protocol of Chronos-2, without additional training or tuning. GIFT-Eval and TIME serve as in-distribution and out-of-distribution benchmarks, respectively. The univariate-trained prompts improve both metrics on both benchmarks.
    }
    \label{tab:multivariate}
    \small
    \setlength{\tabcolsep}{3pt}
    \renewcommand{\arraystretch}{1.1}

    \begin{subtable}[t]{0.48\textwidth}
        \centering
        \caption{In-distribution results on GIFT-Eval}
        \label{tab:multivariate-gift}
        \begin{tabular*}{\linewidth}{@{\extracolsep{\fill}}lrrrr@{}}
            \toprule
            & \multicolumn{2}{c}{Context 4,096}
            & \multicolumn{2}{c}{Context 8,192} \\
            \cmidrule(lr){2-3}
            \cmidrule(lr){4-5}
            Method
            & MASE $\downarrow$ & CRPS $\downarrow$
            & MASE $\downarrow$ & CRPS $\downarrow$ \\
            \midrule
            Zero-shot
            & 0.701 & 0.488 & 0.698 & 0.485 \\
            \method
            & \textbf{0.695} & \textbf{0.478}
            & \textbf{0.690} & \textbf{0.473} \\
            \bottomrule
        \end{tabular*}
    \end{subtable}
    \hfill
    \begin{subtable}[t]{0.48\textwidth}
        \centering
        \caption{Out-of-distribution results on TIME}
        \label{tab:multivariate-time}
        \begin{tabular*}{\linewidth}{@{\extracolsep{\fill}}lrrrr@{}}
            \toprule
            & \multicolumn{2}{c}{Context 4,096}
            & \multicolumn{2}{c}{Context 8,192} \\
            \cmidrule(lr){2-3}
            \cmidrule(lr){4-5}
            Method
            & MASE $\downarrow$ & CRPS $\downarrow$
            & MASE $\downarrow$ & CRPS $\downarrow$ \\
            \midrule
            Zero-shot
            & 0.659 & 0.555 & 0.663 & 0.557 \\
            \method
            & \textbf{0.655} & \textbf{0.553}
            & \textbf{0.658} & \textbf{0.554} \\
            \bottomrule
        \end{tabular*}
    \end{subtable}
\end{table}

\textbf{Multivariate forecasting.}
We further extend the evaluation of \method, trained under the univariate setting, to multivariate forecasting on GIFT-Eval and TIME using the official inference protocol of Chronos-2, without additional training or tuning.
As shown in Table~\ref{tab:multivariate}, \method~improves both MASE and CRPS at both context lengths on GIFT-Eval.
It also yields modest improvements over zero-shot forecasting on the out-of-distribution benchmark TIME, further supporting its generalization ability.
Moreover, \method~at 4,096 steps outperforms zero-shot Chronos-2 at 8,192 steps on both metrics on both benchmarks.
These results show that latent prompts learned under the univariate setting remain effective as context surrogates for multivariate forecasting.

\begin{figure}[t]
\centering
\includegraphics[width=\linewidth]{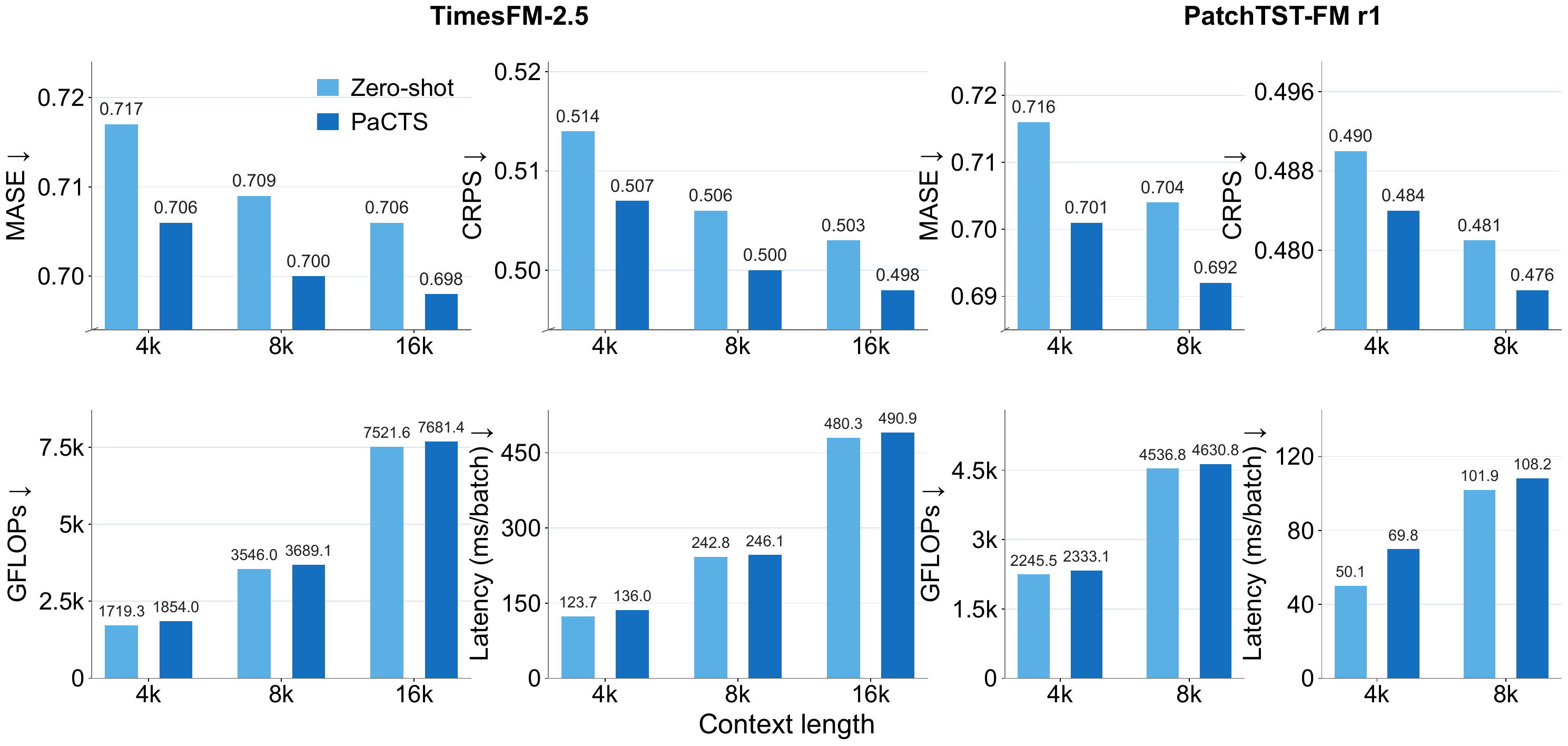}
\vspace{-6mm}
\caption{
    \textbf{Forecasting performance (top) and computational costs (bottom) across backbones} on GIFT-Eval.
    Lower values are better for all metrics.
    As with Chronos-2, shorter-context \method~achieves competitive or better performance than longer-context zero-shot baselines, with substantially lower computational costs.
}
\label{fig:across-backbones}
\end{figure}
\textbf{Results across backbones.}
We further evaluate \method~on encoder-only (PatchTST-FM \citep{wen2026revisiting}) and decoder-only (TimesFM-2.5 \citep{das2024timesfm}) backbones.
As shown in Figure~\ref{fig:across-backbones}, \method~reduces MASE and CRPS at every evaluated context length, including contexts of up to 16,384 steps for TimesFM-2.5.
Across both backbones, shorter-context \method~achieves lower MASE and competitive CRPS compared with zero-shot inference using double context length.
Against zero-shot inference at roughly twice the context length,
\method~at 4,096 steps lowers MASE by 0.43\% on PatchTST-FM
and 0.42\% on TimesFM-2.5, using around 51\% and 52\% of the GFLOPs,
respectively.
On TimesFM-2.5, \method~at 8,192 steps further improves MASE and CRPS over zero-shot inference at 16,384 steps by 0.85\% and 0.60\%, using only 49\% of the GFLOPs.
The efficiency results in the second row show that \method~adds less than 8\% FLOPs at the same context length, while using only about half FLOPs of zero-shot inference with double context length.
The results demonstrate the applicability of \textit{latent prompts as context surrogates} across backbones.
More detailed numbers are provided in Appendix \ref{app:exp}.

\section{Conclusion}
We present a lightweight approach that uses learned input-space prompts as context surrogates for frozen time-series foundation models, improving forecasting without explicitly processing substantially longer histories or updating backbone parameters. Across GIFT-Eval and TIME, the approach provides consistent gains, favorable accuracy--compute trade-offs, and better out-of-distribution transfer than weight-space adaptation, while also extending across multiple TSFM architectures and to multivariate inference. Our current study nevertheless has several limitations: the prompts are trained only under the univariate setting, leaving multivariate prompt training unexplored, and their ability to replace additional context depends on the context-scaling behavior of the underlying backbone. Future work can investigate prompt learning directly in multivariate settings and more general context-surrogate mechanisms across broader TSFM architectures.

\bibliography{iclr2027_conference}
\bibliographystyle{iclr2027_conference}

\newpage
\appendix
\section*{Appendix}

\section{Datasets and Evaluation Protocol}
\label{app:data}

\paragraph{Training and test configurations.}
GIFT-Eval \citep{aksu2024gift} comprises 35 datasets in the setting considered here. Prompt
training pools 48 dataset--frequency tasks, while the main evaluation aggregates 97 dataset--frequency--horizon configurations. 
A configuration, rather than a dataset, is the unit of metric aggregation. 
For each backbone and context-length setting, we train one prompt module on the pooled training tasks and test it for every corresponding test configuration, without dataset-specific fine-tuning.

\paragraph{Temporal boundaries.}
The data loader first withholds the test portion of each series. To prevent overlap with the official evaluation windows on short series, the available
prefix for a series of length $T$ is truncated at
\begin{equation}
T_{\mathrm{avail}}=
\min\!\left(\lfloor 0.9T\rfloor,\,
T-\ell_{\mathrm{test}}\right),
\end{equation}
where $\ell_{\mathrm{test}}$ is the length covered by that task's official
test windows. 
The first $\lfloor0.89T_{\mathrm{avail}}\rfloor$ observations are available to the training sampler, and the remainder of the available prefix supplies validation targets. 
Validation contexts may include earlier training observations, but validation forecast targets begin after the training boundary. This construction is approximately an 80/10/10 temporal split on long series and uses a stricter cutoff where a proportional split would intrude into an official test window.

\paragraph{Transfer benchmark.}
TIME \citep{qiao2026s} contains 98 forecasting tasks from 50 datasets with no dataset overlap with GIFT-Eval. We apply the GIFT-trained prompt module directly to TIME without updating either the module or the backbone. 
The same transfer question is evaluated for the other adaptation baselines. 
Therefore, GIFT-Eval test results measure in-distribution performance while TIME evaluates transferring to out-of-distribution datasets.

\paragraph{Metrics.}
We follow the GIFT-Eval evaluation protocol and report mean absolute
scaled error (MASE) and CRPS \citep{aksu2024gift}. Let
$\mathcal I_c$ contain all valid forecast positions across windows and
variates in test set $c$. For window $i$, let $x_{i,1:L_i}$ be its
observed history, $s_i$ its seasonal period, $y_{it}$ the target, and
$\hat y_{it}^{(q)}$ the predicted $q$-quantile. The seasonal scale and
test-set MASE are
\begin{align}
d_i =
\frac{1}{L_i-s_i}
\sum_{t=s_i+1}^{L_i}|x_{it}-x_{i,t-s_i}|,\\
\operatorname{MASE}_c =
\frac{1}{|\mathcal I_c|}
\sum_{(i,t)\in\mathcal I_c}
\frac{|y_{it}-\hat y_{it}^{(0.5)}|}{d_i}.
\end{align}
The metric labeled CRPS by GIFT-Eval is the mean weighted sum quantile
loss over $\mathcal Q=\{0.1,0.2,\ldots,0.9\}$. With
$\rho_q(u)=u\bigl(q-\mathbf{1}\{u<0\}\bigr)$, it is computed as
\begin{equation}
\operatorname{CRPS}_c =
\frac{1}{|\mathcal Q|}
\sum_{q\in\mathcal Q}
\frac{
2\sum_{(i,t)\in\mathcal I_c}
\rho_q(y_{it}-\hat y_{it}^{(q)})
}{
\sum_{(i,t)\in\mathcal I_c}|y_{it}|
}.
\end{equation}
Thus, quantile losses and absolute targets are pooled across valid
positions within each test set before their ratio is taken.

Following the GIFT-Eval leaderboard, we divide each test-set score by
the Seasonal Naive score computed using the same metric, then take the
geometric mean over all 97 test sets $\mathcal C$:
\begin{equation}
G_m(\mathcal C)=
\exp\!\left[
\frac{1}{|\mathcal C|}
\sum_{c\in\mathcal C}
\log\!\left(\frac{m_c}{m_c^{\mathrm{SN}}}\right)
\right],
\qquad m\in\{\operatorname{MASE},\operatorname{CRPS}\}.
\end{equation}
We use the same MASE and quantile-based CRPS metrics for TIME, following its official window-level aggregation protocol \citep{qiao2026s}.

\section{Statistics of Prompt Construction}
\label{sec:stats}

The generator computes ten statistics from the normalized visible input, using the observation mask to exclude missing or padded values.
Table~\ref{tab:app-stats} lists the implemented features. The same extractor is applied to the full context and to each valid segment.
The statistics include changes across time and lag-one dependence. 
In the segment branch, the configured seasonal period can affect segment granularity, and the ordered segment descriptors retain local temporal variation.
All statistics are computed non-parametrically and are invariant to the location and scale of the original series.

\begin{table*}[t]
\centering
\small
\caption{\textbf{Ten input statistics used by the prompt generator.}}
\label{tab:app-stats}
\begin{tabular}{cll}
\toprule
No. & Feature & Implementation \\
\midrule
1 & Mean & Mean of observed values \\
2 & Standard deviation & Clipped above at 10 \\
3 & Minimum & Clipped to $[-10,10]$ \\
4 & Maximum & Clipped to $[-10,10]$ \\
5 & Half-window change & Second-half mean minus first-half mean; $[-5,5]$ \\
6 & Mean first difference & Consecutive observed pairs; $[-5,5]$ \\
7 & Std. first difference & Consecutive observed pairs; clipped at 10 \\
8 & Lag-one autocorrelation & Clipped to $[-1,1]$ \\
9 & Observed fraction & Observed length divided by input width \\
10 & Log length & $\log(1+n_{\mathrm{obs}})/10$ \\
\bottomrule
\end{tabular}
\end{table*}

\section{Training and Adaptation Baselines}
\label{app:training}

\paragraph{Prompt optimization.}
The default adaptive generator has rank $r=4$, and the prompt module uses $M=10$ tokens and up to $S=16$ segments. 
We use temperature-based task sampling with probability proportional to $N_k^{0.2}$, where $N_k$ is the number of univariate series entries retained for task $k$.
Within each sampled entry, we draw eight training windows when valid prediction origins are available.
For each task, we retain at most the number of source-series records specified in the table before expanding multivariate records into univariate entries.
We optimize the parameters with AdamW at learning rate $10^{-3}$ and select checkpoints by validation loss. 
The reported prompt runs use an effective batch size of 48 for at most 12 epochs.
The quantile loss is normalized by each sample's number of valid forecast positions before averaging across samples, so a shorter valid horizon is not down-weighted by padding. For \method, all backbone parameters remain frozen.

For segment extraction, the reference width is $W=\max(1,\lfloor C/S\rfloor)$, where $C$ is the configured context length. In the adaptive-horizon setting, the minimum segment length is $\min(W,\max(L_{\min},p))$, where $L_{\min}=48$ and $p$ is the configured seasonal period if greater than one, or the forecast horizon otherwise.
The actual segment length also depends on the available history and cannot exceed its observed length. The cross-attention module uses width 64 and four heads.

We prepend the prompts before the Transformer layers while retaining the original time-series tokens and their zero-shot positions. 
The prompt tokens are assigned position ID of 0.
All three backbones use the same statistics-to-basis MLP with two hidden layers of width 32 and the same segment-refinement design.
The rank-four basis is initialized from a zero-mean Gaussian with standard deviation $10^{-3}$, and the shared prompts are initialized from the mean visible input-token embedding of the corresponding backbone's training data.

\begin{table*}[t]
\centering
\small
\caption{\textbf{Chronos-2 adaptation baseline settings.} Every row uses the official fit pipeline, 5,000 optimizer steps, and batch size 256.}
\label{tab:app-baseline-protocol}
\begin{tabular}{lll}
\toprule
Method & Updated parameters & Learning rate \\
\midrule
Full fine-tuning & All backbone parameters & $10^{-6}$ \\
LoRA & Rank 8, $\alpha=16$; attention and output layers & $10^{-5}$ \\
LayerNorm-only & Layer-normalization parameters & $10^{-3}$ \\
BitFit & Selected bias parameters & $10^{-3}$ \\
Linear probe & Output patch embedding & $10^{-3}$ \\
\bottomrule
\end{tabular}
\end{table*}

\paragraph{Baseline optimization.}
On Chronos-2, full fine-tuning and LoRA follow the official
\texttt{Chronos2Pipeline.fit()} training procedure. LayerNorm-only tuning, BitFit, and linear probing use the same pipeline, restricting the optimizer to the selected parameter subset. All five baselines use an effective batch size of 256 and train for 5,000 optimizer steps. Validation is performed every 100 steps, and the best validation checkpoint is selected.
Table~\ref{tab:app-baseline-protocol} lists the updated parameter subsets and learning rates.

\section{Additional Experimental Results}
\label{app:exp}

\paragraph{Detailed results of in-distribution and out-of-distribution generalization comparisons.}
Table~\ref{tab:app-transfer} lists the absolute scores underlying the GIFT-Eval and TIME comparison. Every adapted row uses a checkpoint trained on GIFT-Eval. The zero-shot rows evaluate pretrained Chronos-2 without adaptation.
    
\begin{table*}[t]
\centering
\small
\caption{\textbf{Detailed results of in-distribution and out-of-distribution generalization comparisons.}  \method~achieves the best in-distribution performance on GIFT-Eval and is the only method that improves both metrics over zero-shot inference on the out-of-distribution TIME benchmark.}
\label{tab:app-transfer}
\resizebox{\textwidth}{!}{%
\begin{tabular}{llrrrrrrrr}
\toprule
& & \multicolumn{4}{c}{GIFT-Eval}
& \multicolumn{4}{c}{TIME} \\
\cmidrule(lr){3-6}\cmidrule(lr){7-10}
Method &  Context steps & MASE & $\Delta$ (\%) & CRPS & $\Delta$ (\%)
       & MASE & $\Delta$ (\%) & CRPS & $\Delta$ (\%) \\
\midrule
Zero-shot &  4,096        & 0.708 &  - & 0.496 &  - & 0.664 &  - & 0.558 &  - \\
Zero-shot &  {8,192}        & 0.704 & -0.56 & 0.491 & -1.01 & 0.666 & +0.30 & 0.559 &  +0.18 \\
\midrule
Full fine-tuning &  4,096 & 0.703 & -0.71 & 0.487 & -1.81 & 0.665 & +0.15 & 0.560 & +0.36 \\
LoRA &  4,096             & 0.703 & -0.71 & 0.493 & -0.60 & 0.666 & +0.30 & 0.560 & +0.36 \\
LayerNorm-only &  4,096   & 0.705 & -0.42 & 0.488 & -1.61 & 0.666 & +0.30 & 0.559 & +0.18 \\
BitFit &  4,096           & 0.708 &  0.00 & 0.489 & -1.41 & 0.671 & +1.05 & 0.564 & +1.08 \\
Linear probe &  4,096     & 0.710 & +0.28 & 0.491 & -1.01 & 0.672 & +1.20 & 0.565 & +1.25 \\
\rowcolor{lightblue}
\method &  4,096          & \textbf{0.697} & \textbf{-1.55} & \textbf{0.480} & \textbf{-3.23} & \textbf{0.661} & \textbf{-0.45} & \textbf{0.555} & \textbf{-0.54} \\
\bottomrule
\end{tabular}}
\end{table*}

\paragraph{Effects of history length.}
Table~\ref{tab:app-history} reports detailed results for all 97
GIFT-Eval configurations, grouped by the median available history length of their dataset. Each score is the seasonal-naive-normalized geometric mean within the indicated group. 
As discussed in Figure~\ref{fig:history-horizon}, \method~ improves both metrics and outperforms adaptation baselines and zero-shot inference with double the context length in both groups, supporting the effectiveness of latent prompts as context surrogates.

\begin{table*}[t]
\centering
\small
\caption{\textbf{Detailed comparisons of different history lengths on GIFT-Eval.}}
\label{tab:app-history}
\begin{tabular}{llrrrr}
\toprule
& & \multicolumn{2}{c}{History $<4,096$ ($n=38$)}
& \multicolumn{2}{c}{History $\geq4,096$ ($n=59$)} \\
Method & Context steps  & MASE & CRPS & MASE & CRPS \\
\midrule
Zero-shot & 4,096        & 0.700 & 0.533 & 0.712 & 0.474 \\
Zero-shot & 8,192        & 0.700 & 0.533 & 0.706 & 0.466 \\
\midrule
LoRA & 4,096             & 0.690 & 0.524 & 0.711 & 0.474 \\
Full fine-tuning & 4,096 & 0.690 & 0.519 & 0.711 & 0.468 \\
\rowcolor{lightblue}
\method & 4,096          & \textbf{0.685} & \textbf{0.511} & \textbf{0.705} & \textbf{0.461} \\
\bottomrule
\end{tabular}
\end{table*}

\paragraph{Effects of Forecast horizon.}
Table~\ref{tab:app-horizon} partitions the same 97 configurations by forecast horizon and reports detailed scores.
The groups are defined following the official definition of GIFT-Eval \citep{aksu2024gift}.
Each score is the seasonal-naive-normalized geometric mean within the indicated group. 
As discussed in Figure~\ref{fig:history-horizon}, \method~ outperforms both adaptation baselines and double-context inference again on both metrics across all three horizon groups, which demonstrates consistent benefits across different forecasting lengths.

\paragraph{Results across backbones.}
We also provide the detailed results based on PatchTSTFM-r1 \citep{wen2026revisiting} and TimesFM-2.5 \citep{das2024timesfm} in
Tables~\ref{tab:app-r1-performance} and
\ref{tab:app-tfm25-performance}, respectively. 
Compared with the zero-shot pretrained baseline, \method~achieves better performance on both backbones across both metrics at each context length on GIFT-Eval.

\begin{table*}[!t]
\centering
\small
\caption{\textbf{Detailed comparisons of different forecast horizons on GIFT-Eval.}}
\label{tab:app-horizon}
\begin{tabular}{llrrrrrr}
\toprule
& & \multicolumn{2}{c}{Short ($n=55$)}
& \multicolumn{2}{c}{Medium ($n=21$)}
& \multicolumn{2}{c}{Long ($n=21$)} \\
Method & Context steps & MASE & CRPS & MASE & CRPS & MASE & CRPS \\
\midrule
Zero-shot & 4,096        & 0.676 & 0.506 & 0.740 & 0.485 & 0.763 & 0.483 \\
Zero-shot & 8,192        & 0.675 & 0.506 & 0.729 & 0.468 & 0.757 & 0.477 \\
\midrule
Full fine-tuning & 4,096 & 0.670 & 0.497 & 0.736 & 0.476 & 0.760 & 0.474 \\
LoRA & 4,096             & 0.670 & 0.501 & 0.734 & 0.481 & 0.761 & 0.484 \\
\rowcolor{lightblue}
\method & 4,096          & \textbf{0.667} & \textbf{0.491} & \textbf{0.728} & \textbf{0.466} & \textbf{0.749} & \textbf{0.463} \\
\bottomrule
\end{tabular}
\end{table*}

\begin{table*}[!t]
\centering
\small
\caption{\textbf{GIFT-Eval performance of PatchTST-FM at different context lengths.} The 8,192-step model window reserves a 128-step masked forecast span for the 96-step horizon, leaving 8,064 context steps.}
\label{tab:app-r1-performance}
\begin{tabular}{lrrrrrr}
\toprule
& \multicolumn{3}{c}{MASE$\downarrow$}
& \multicolumn{3}{c}{CRPS$\downarrow$} \\
\cmidrule(lr){2-4}
\cmidrule(lr){5-7}
Context & Zero-shot & \method & $\Delta$ (\%)
        & Zero-shot & \method & $\Delta$ (\%) \\
\midrule
4,096 & 0.716 & \textbf{0.701} & $-2.09$
      & 0.490 & \textbf{0.484} & $-1.22$ \\
8,064 & 0.704 & \textbf{0.692} & $-1.70$
      & 0.481 & \textbf{0.476} & $-1.04$ \\
\bottomrule
\end{tabular}
\end{table*}

\begin{table*}[!t]
\centering
\small
\caption{\textbf{GIFT-Eval performance of TimesFM-2.5 at matched context
lengths.}}
\label{tab:app-tfm25-performance}
\begin{tabular}{lrrrrrr}
\toprule
& \multicolumn{3}{c}{MASE$\downarrow$}
& \multicolumn{3}{c}{CRPS$\downarrow$} \\
\cmidrule(lr){2-4}
\cmidrule(lr){5-7}
Context & Zero-shot & \method & $\Delta$ (\%)
        & Zero-shot & \method & $\Delta$ (\%) \\
\midrule
4,096  & 0.717 & \textbf{0.706} & $-1.53$
       & 0.514 & \textbf{0.507} & $-1.36$ \\
8,192  & 0.709 & \textbf{0.700} & $-1.27$
       & 0.506 & \textbf{0.500} & $-1.19$ \\
16,384 & 0.706 & \textbf{0.698} & $-1.13$
       & 0.503 & \textbf{0.498} & $-0.99$ \\
\bottomrule
\end{tabular}
\end{table*}

We also evaluate the efficiency of \method~on the two backbones.
Table \ref{tab:app-backbone-config} shows the configurations and trainable parameters for each backbone, where \method~introduce less than 0.2\% additional parameters for each backbone.
Tables \ref{tab:app-r1-cost}--\ref{tab:app-tfm25-8k-cost} compare \method\ at context $C$ with zero-shot inference at $C$ and $2C$.
The conclusion is similar as that for Chronos-2.
Relative to zero-shot inference at about doubled context, \method\ reduces GFLOPs by 48.57\% on PatchTST-FM r1 at 4,096 steps, and by 47.72\% and 50.95\% on TimesFM-2.5 at 4,096 and 8,192 steps, respectively, with consistently better MASE and competitive CRPS.
Peak memory and observed latency also decrease in all three comparisons.

\begin{table*}[t]
\centering
\small
\caption{\textbf{Prompt-module configurations in the cross-backbone experiments.}
The trainable fraction is relative to the parameter counts of each frozen backbone.}
\label{tab:app-backbone-config}
\begin{tabular}{lrrrr}
\toprule
Backbone & Width $d$ & Context lengths & Trainable parameters
& Trainable fraction \\
\midrule
Chronos-2 & 768 & 4,096 / 8,192 & 153,286 & 0.128\% \\
PatchTST-FM & 1,024 & 4,096 / 8,064 & 199,110 & 0.077\% \\
TimesFM-2.5 & 1,280 & 4,096 / 8,192 / 16,384 & 244,934 & 0.106\% \\
\bottomrule
\end{tabular}
\end{table*}

\begin{table*}[t]
\centering
\small
\caption{\textbf{GIFT-Eval performance and inference costs of PatchTST-FM r1.}}
\label{tab:app-r1-cost}
\resizebox{\columnwidth}{!}{%
\begin{tabular}{lrrrrrr}
\toprule
Method (context) & MASE$\downarrow$ & CRPS$\downarrow$
& GFLOPs$\downarrow$ & Memory (MiB)$\downarrow$
& Latency (ms)$\downarrow$ & Params (M) \\
\midrule
Zero-shot (4,096) & 0.716 & 0.490 & 2,245.5 & 1,199.4 & 50.1  & 257.896 \\
Zero-shot (8,064) & 0.704 & 0.481 & 4,536.8 & 1,370.8 & 101.9 & 257.896 \\
\rowcolor{lightblue}
\method\ (4,096)  & 0.701 & 0.484 & 2,333.1 & 1,211.2 & 69.8  & 258.095 \\
\midrule
$\Delta$ vs.\ zero-shot (8,064)
& $-0.43\%$ & $+0.62\%$ & $-48.57\%$
& $-11.64\%$ & $-31.50\%$ & $+0.08\%$ \\
\bottomrule
\end{tabular}}
\end{table*}

\begin{table*}[t]
\centering
\small
\caption{\textbf{GIFT-Eval performance and inference costs of TimesFM-2.5 at the 4,096-step comparison point.} }
\label{tab:app-tfm25-4k-cost}
\resizebox{\columnwidth}{!}{%
\begin{tabular}{lrrrrrr}
\toprule
Method (context) & MASE$\downarrow$ & CRPS$\downarrow$
& GFLOPs$\downarrow$ & Memory (MiB)$\downarrow$
& Latency (ms)$\downarrow$ & Params (M) \\
\midrule
Zero-shot (4,096) & 0.717 & 0.514 & 1,719.3  & 1,006.7 & 123.6  & 231.289 \\
Zero-shot (8,192) & 0.709 & 0.506 & 3,546.0 & 1,099.5 & 242.7 & 231.289 \\
\rowcolor{lightblue}
\method\ (4,096)  & 0.706 & 0.507 & 1,854.0  & 1,015.6 & 135.9  & 231.534 \\
\midrule
$\Delta$ vs.\ zero-shot (8,192)
& $-0.42\%$ & $+0.20\%$ & $-47.72\%$
& $-7.63\%$ & $-43.98\%$ & $+0.11\%$ \\
\bottomrule
\end{tabular}}
\end{table*}

\begin{table*}[t]
\centering
\small
\caption{\textbf{GIFT-Eval performance and inference costs of TimesFM-2.5 at longer contexts. }}
\label{tab:app-tfm25-8k-cost}
\resizebox{\columnwidth}{!}{%
\begin{tabular}{lrrrrrr}
\toprule
Method (context) & MASE$\downarrow$ & CRPS$\downarrow$
& GFLOPs$\downarrow$ & Memory (MiB)$\downarrow$
& Latency (ms)$\downarrow$ & Params (M) \\
\midrule
Zero-shot (8,192)  & 0.709 & 0.506 & 3,546.0 & 1,099.5 & 242.7 & 231.289 \\
Zero-shot (16,384) & 0.706 & 0.503 & 7,521.6 & 1,286.7 & 480.3 & 231.289 \\
\rowcolor{lightblue}
\method\ (8,192)   & 0.700 & 0.500 & 3,689.1 & 1,108.8 & 246.0 & 231.534 \\
\midrule
$\Delta$ vs.\ zero-shot (16,384)
& $-0.85\%$ & $-0.60\%$ & $-50.95\%$
& $-13.83\%$ & $-48.77\%$ & $+0.11\%$ \\
\bottomrule
\end{tabular}}
\end{table*}

\end{document}